\documentclass{article}

\usepackage{microtype}
\usepackage{graphicx}
\usepackage{subfigure}
\usepackage{booktabs} % for professional tables

\usepackage{hyperref}

\usepackage[accepted]{icml2025}

\usepackage{amsmath}
\usepackage{amssymb}
\usepackage{mathtools}
\usepackage{amsthm}
\usepackage{csquotes}

\usepackage[capitalize,noabbrev]{cleveref}

\usepackage[binary-units]{siunitx} % For aligning numbers by decimal point in tables
\usepackage{algorithm}
\usepackage{algorithmic}

\usepackage{listings}
\usepackage{caption}
\usepackage[binary-units]{siunitx} % For aligning numbers by decimal point in tables
\usepackage{tabularx} % For tables with adjustable-width columns
\usepackage{makecell} % For multiline cells in tables
\usepackage{graphicx} % Required for \resizebox
\usepackage{subcaption}

\theoremstyle{plain}

\theoremstyle{definition}

\theoremstyle{remark}

\usepackage[textsize=tiny]{todonotes}

\icmltitlerunning{ELMO : Efficiency via Low-precision and Peak Memory Optimization in Large Output Spaces}

\begin{document}

\twocolumn[
% \icmltitle{SlimXML: Scaling with Low-precision and peak Memory optimization for eXtreme MultiLabel classification}
\icmltitle{ELMO : Efficiency via Low-precision and Peak Memory Optimization \\ in Large Output Spaces}

% Alternate title: Scaling Extreme Classification: Low-Precision and Memory-Efficient Training for Large Output Spaces



\icmlsetsymbol{equal}{*}

\begin{icmlauthorlist}
\icmlauthor{Jinbin Zhang}{equal,yyy}
\icmlauthor{Nasib Ullah}{equal,yyy}
\icmlauthor{Erik Schultheis}{yyy,comp}
\icmlauthor{Rohit  Babbar}{sch}
%\icmlauthor{Firstname8 Lastname8}{sch}
%\icmlauthor{Firstname8 Lastname8}{yyy,comp}
\end{icmlauthorlist}

\icmlaffiliation{yyy}{Department of Computer Science, Aalto University, Espoo, Finland}
\icmlaffiliation{comp}{IST Austria}
\icmlaffiliation{sch}{Department of Computer Science, University of Bath, Bath, UK}

\icmlcorrespondingauthor{Jinbin Zhang}{jinbin.zhang@aalto.fi}
%\icmlcorrespondingauthor{Firstname2 Lastname2}{first2.last2@www.uk}

\icmlkeywords{Machine Learning, ICML}

\vskip 0.3in
]



\printAffiliationsAndNotice{\icmlEqualContribution} % otherwise use the standard text.

\begin{abstract}
Large output spaces, also referred to as Extreme multilabel classification (XMC), is a setting that arises, e.g., in large-scale tagging and product-to-product recommendation, and is characterized by the number of labels ranging from hundreds of thousands to millions. This means that the linear classification head, usually only a tiny fraction of the overall model, turns into the main driver for compute and memory demand. Current state-of-the-art XMC methods predominantly rely on \texttt{FP16-FP32} mixed-precision training, which we show can be unstable, and inefficient in terms of memory usage and computational overhead. Meanwhile, existing low-precision methods typically retain higher precision for the classification layer. In this work, we propose ELMO, a pure low-precision training framework for XMC models using \texttt{BFloat16} and \texttt{Float8} data types. By leveraging Kahan summation and stochastic rounding, we demonstrate that XMC models can be effectively trained entirely in \texttt{\texttt{Float8}}, without relying on single-precision master weights or tensor scaling. Low-precision training, combined with our proposed memory optimizations---gradient fusion and chunking---enables significant reductions in GPU memory usage. For example, we train a 3-million-label XMC model with only \SI{6.6}{\gibi\byte} of GPU memory, compared to the \SI{39.7}{\gibi\byte} required by the optimized SOTA method, Renee \cite{jain2023renee} without compromising accuracy.
%
%. Notably, our efficiency gains do not come at the expense of performance: ELMO competes or outperforms existing baselines on public datasets, achieving the non-trivial goal of improving performance under purely low-precision constraints.  
Code available at \href{https://github.com/xmc-aalto/elmo}{\texttt{https://github.com/xmc-aalto/elmo}}.
\end{abstract}

\section{Introduction}
\label{intro}
Large output spaces, also referred to as Extreme multilabel classification (XMC) \cite{Bhatia16,babbar2017dismec,prabhu2018parabel} refers to the task of predicting a sparse subset of relevant labels from an exceedingly large set, often ranging from hundreds of thousands to millions of potential classes. XMC has gained prominence due to its applicability in real-world scenarios such as product recommendations, Wikipedia tagging, and matching search queries to advertisements. From the perspective of standard deep learning literature, this problem appears solvable using an encoder such as a CNN \cite{hu2014convolutional}, LSTM \cite{hochreiter1997long,chung2014empirical}, or more commonly, a Transformer \cite{vaswani2017attention,devlin2018bert} fine-tuned with a linear output layer in what is typically referred to as an end-to-end approach. 
However, contrary to other domains, where a transformer model such as BERT \cite{devlin2018bert}
would account for a vast majority of all model parameters, in XMC, it is the classifier layer that becomes the bottleneck.
For example, with an embedding dimension of 768, and three million labels, classifier weights alone would consume
approximately \SI{8}{\gibi\byte} of memory. 
When accounting for gradients and optimizer states \cite{kingma2014adam}, 
the memory footprint expands to around \SI{32} {\gibi\byte}. 
Furthermore, loss computation in the last layer of the network, involving billions of parameters for larger datasets, also entails an enormous computational challenge.

Renee \citep{jain2023renee} shows that full end-to-end training can be feasible with appropriate memory and computational optimizations in training the model, and results in classification performance superior to the approaches employing negative sampling \cite{jiang2021lightxml,zhang2021fast,kharbanda2022cascadexml}. 
It exploits the fact that, particularly for neural networks in large output spaces, the loss and gradient computation using automatic differentiation engines involves maintaining large buffers for intermediate states.
%such as logits, and target label vectors. 
Therefore, as long as one can directly compute the gradients required for the backward pass, explicitly computing the loss can be forgone, hence avoiding to materialize the memory allocations for the intermediate variables altogether. 
In this way, Renee achieves a significant reduction in activation memory required for the classification layer. 

Despite the model optimizations in Renee, several memory bottlenecks remain. 
\textit{Firstly}, Renee employs mixed precision \cite{micikevicius2018mixed} training for the encoder, coupled with standard gradient scaling, which requires maintaining a full precision copy of the parameters. Furthermore, this approach mandates that input gradients be kept in full precision, leading to the classifier layer’s gradients also being cast to full precision, which further inflates memory usage. 
\textit{Secondly}, through memory snapshot analysis \ref{fig:pmo}, we observe that in Renee, the order of execution in the computation graph causes memory-intensive operations to accumulate at a single point in time, leading to excessive peak memory demand.
\textit{Thirdly}, neither Renee, nor any of the label shortlisting approaches, result in a reduction in memory requirements for the classification layer weights. 

To address these challenges, we first move from mixed precision to pure 16-bit training, achieving concurrent reductions in both memory and computation time. We adopt \texttt{BFloat16} (\texttt{BF16}) for gradients, leveraging its extended range to mitigate the training instability caused by potential overflows. To compensate for precision loss, we employ Kahan summation \cite{kahan} for the encoder optimizer and stochastic rounding \cite{zhang2018training} for the classifier optimizer, compensating for inaccuracies and rounding bias during parameter updates. Going further, we demonstrate that classifiers can be trained in pure \texttt{Float8} (\texttt{FP8}) without scaling or mixed-precision training, as long as gradients remain in \texttt{BF16}. Finally, by integrating a \texttt{FP8} encoder from \texttt{torchao}, we achieve a nearly pure \texttt{FP8} (excluding gradients of the transformer backbone) training pipeline for XMC tasks.

Adopting pure \texttt{BF16} and \texttt{FP8} training reduces parameter memory requirements by 50–75\% relative to full precision. 
%However, we show that standard mixed-precision training can impose severe memory overheads in XMC. 
In order to address memory accumulation in Renee, we reorganize the computation flow and decouple encoder and classifier updates, leading to more evenly distributed memory allocations throughout each training iteration and reducing peak memory usage. Furthermore, our chunking strategy for classifier updates curtails transient memory demands, and by fusing classifier gradient computation with the optimizer step in a custom Triton kernel, we effectively eliminate the need to store classifier gradients in memory. With these optimizations, the proposed method, ELMO, requires only \SI{10.3}{\gibi\byte}\footnote{$1GB \approx 0.93GiB$} (\texttt{BF16}) or \SI{6.6}{\gibi\byte} (\texttt{FP8}) of memory for a 3-million-label model, significantly lower than \SI{39.7}{\gibi\byte} necessitated by Renee.

We evaluate our low-precision training method on datasets of varying labels' sizes, demonstrating results comparable to Renee. To further assess the efficiency of our approach, we derived a new dataset with 8.6M labels from the DBLP-Citation-network V14 dataset~\cite{Tang:08KDD, Tang:10TKDD, Tang:11ML, Tang:12TKDE, Tang:07ICDM, sinha2015overview}. On this larger dataset, our low-precision training method demonstrates significant memory and computational savings. We believe that low-precision training will become a standard in the XMC field, given the growing demand for handling datasets with an immense number of labels.

To summarize, this paper makes the following contributions: (1) We introduce a purely low-precision training approach using \texttt{BF16} and \texttt{FP8} for models with large classification layers. (2) Combined with peak memory optimizations, this reduces memory usage by 4x–6x for a 3-million-label dataset; (3) Apart from efficiency gains, we compete with existing XMC baselines on most public datasets, illustrating that purely low-precision training can preserve similar performance; (4) We introduce LF-Paper2Keywords-8.6M, an 8.6-million-label dataset that, to our knowledge, is now the largest publicly available XMC benchmark.

\section{Related Work}

We presented discussion of Renee, the most relevant XMC method for our work, in the previous section. 
%Due to space constraints, we refer the reader to Appendix \ref{related_work_xmc} for further discussion on various classes of other XMC methods. 
Next, we provide a brief overview of different categories of other XMC methods and low-precision training.

\noindent \textbf{Extreme Classification.} Initial approaches in extreme classification utilized linear classifiers with bag-of-words and TF-IDF features \citep{babbar2017dismec,babbar2019data}. 
As the computational cost of these methods scales linearly with the number of labels, tree-based methods were introduced to reduce computation complexity to logarithmic scale with respect to label count \citep{prabhu2018parabel,khandagale2020bonsai,wydmuch2018no,jasinska2021online}. 
Subsequent advancements incorporated task-specific feature learning using deep encoders atop label trees, under the assumption that joint training of both encoder and classifier layers would be computationally expensive without some form of label shortlisting for negative sampling when evaluating the loss function in the last layer \cite{you2019attentionxml,jiang2021lightxml,kharbanda2022cascadexml,zhang2021fast, kharbanda2023inceptionxml}. While tree-based methods provided a negative sampling mechanism, alternative approaches employed nearest-neighbor approach for negative  sampling and multi-stage training to first train the encoder, followed by the classifier \cite{dahiya2021siamesexml,dahiya2023ngame,mittal2021eclare,dahiya2023deep, kharbanda2025unidec}. DEXML \cite{gupta2024dualencoders} eliminates the classifier using dual-encoder training with all-negative labels, but at the cost of increased compute and memory usage.
More recently, Renee \cite{jain2023renee} demonstrated the feasibility of a fully end-to-end approach while optimizing memory consumption, revealing its potential to outperform traditional sampling-based methods. 
Our work advances this end-to-end approach, focusing on enhancing computational and memory efficiency by leveraging recent developments in low-precision training.

\noindent \textbf{Low Precision Training.} With the success of foundation models \cite{brown2020language,achiam2023gpt,zhang2022opt,touvron2023llama}, there has been a strong push toward scaling both model size and training data. Nevertheless, this scalability is limited by memory and computational constraints, where low-precision training provides a notable benefit. \citet{micikevicius2018mixed} introduced a method for training in mixed precision (\texttt{FP16-FP32}) by maintaining a master copy in full precision and using loss scaling to counteract gradient underflow. The restricted range of \texttt{FP16} may cause instability in large models \cite{rae2021scaling,zeng2022glm}, necessitating a transition to \texttt{BF16-FP32} mixed precision \cite{rae2021scaling,smith2022using} for enhanced stability. An alternative to mixed-precision training, \citet{zamirai2020revisiting} employs pure \texttt{BF16} with methods such as Kahan summation \cite{kahan} and stochastic rounding \cite{forsythe1950round,croci2022stochastic} to reduce rounding errors, while COLLAGE \cite{yucollage} utilizes pure \texttt{BF16} with a multi-component floating-point representation \cite{yu2022mctensor} for enhanced precision. The \texttt{FP8} format is a promising next step for further reduction in precision, with early successes in using tensor scaling, as demonstrated by \citet{micikevicius2022fp8}, and hybrid formats \cite{sun2019hybrid} for weights and gradients with a different exponent bias. %\citet{dettmers2022bit} and \citet{} reduces the precision of Adam optimizer states to 8-bit and 4-bit respectively, resulting in substantial memory savings. 
Currently, \texttt{FP8} support \cite{torchao} is limited primarily to \texttt{matmul()} operations. Towards this \citet{peng2023fp8} introduces a comprehensive FP8 training suite, including an \texttt{FP8} optimizer, communication, and distributed training capabilities. Another line of work focuses on reducing memory in Adam optimizer states \cite{dettmers2022bit, zhaogalore} or using stateless optimizer \cite{lv2024full}. 
In the XMC setup, recent works have successfully achieved compression of the last layer using dynamic sparse training \cite{schultheis2023towards, ullahnavigating}. As an orthogonal approach, in this paper, we focus on low-precision, specifically \texttt{FP8} training for XMC models, where the classifier can benefit significantly from the memory and computation efficiency of \texttt{FP8} representation.

\section{Problem setup and preliminaries}

\noindent \textbf{Problem setup} For a multi-label dataset with \(N\) training samples, \(\mathcal{D} = \{(x_i, P_i)_{i=1}^{N}\}\); \(L\) as the total number of labels, and \(P_i \subset [L]\) denotes a subset of relevant labels associated with \(x_i\in \mathcal{X}\) such that $|P_i| \ll L,  \forall i$.
Typically, the instances are text based, such as the contents of an article on Wikipedia or the description of a product
on Amazon with labels being Wikipedia categories and frequently bought together products, respectively ~\citep{Bhatia16}.
%Traditional XMC methods used to handle labels the same way as is typically done in other fields, as featureless integers.

% \noindent Compensation and Kahan Summation:
\noindent \textbf{Floating-point formats} The standard binary floating-point representation, defined in IEEE 754 \cite{IEEE754-2019}, is specified by the number of exponent bits $E$, mantissa bits $M$.
the format includes special bit patterns for values like \texttt{infinities}, \texttt{NaNs}, and \texttt{subnormals}, which represent very small magnitudes near zero. The common 32-bit floating-point (\texttt{FP32}) format uses $23$ mantissa bits and $8$ exponent bits. Lower-precision formats, like \texttt{FP16} (half-precision) and \texttt{BF16} (\texttt{BF16}), reduce mantissa and/or exponent bits. \texttt{BF16}, for example, retains the 8 exponent bits of \texttt{FP32} but has fewer mantissa bits, offering a similar dynamic range with reduced precision. Recently, \texttt{FP8} formats have also been proposed, with $4$ or $5$ exponent bits and $3$ or $2$ mantissa bits (referred to as \texttt{E4M3} and \texttt{E5M2} respectively), using the IEEE 754 structure but differ in how they handle special values.

The reduction in exponent and mantissa bits in lower-precision formats introduces \emph{quantization errors}. \emph{Clipping error} occurs when the exponent range is insufficient, causing values outside the representable range to be ``clipped" to the nearest maximum or minimum representable value. Meanwhile, \emph{rounding error} arises from having fewer mantissa bits, which forces values to be rounded to the nearest representable value within the available precision. As a result, small weight updates are canceled due to nearest rounding \cite{zamirai2020revisiting}.

\noindent \textbf{Stochastic Rounding and Kahan summation.} For any finite subset $F$ of the reals $\mathbb{R}$, the \emph{stochastic rounding}~\citep{forsythe1950round,croci2022stochastic, zamirai2020revisiting} of $x \in \mathbb{R}$, is defined as follows :
\begin{equation}
    \operatorname{SR}(x) = 
    \begin{cases}
        \left\lceil x \right\rceil & \mbox{with probability $p(x)$}\\
        \left\lfloor x \right\rfloor & \mbox{with probability $1 - p(x)$}
    \end{cases}
    \label{eq:sr}
\end{equation}
where $\left\lceil x \right\rceil = \min \{z \in F : z \geq x\}$,  $\left\lfloor x \right\rfloor = \max \{z \in F : z \leq x\}$ denote rounding up and down, respectively.
The probability is based on the distance $p(x) = \frac{x - \left\lfloor x \right\rfloor}{\left\lceil x \right\rceil - \left\lfloor x \right\rfloor}$.
While stochastic rounding does not make any \emph{individual} rounding operation more accurate, its result is an unbiased estimate
of the true number. Thus stochastic rounding can prevent the catastrophic accumulation of rounding errors, e.g., when adding many small numbers sequentially to one large number, as is likely to happen when adding small gradient updates to the network's weights.

An alternative is to employ \emph{Kahan summation}~\citep{kahan}. In this case, an additional buffer keeps track of the rounding error, and is used
to correct subsequent additions:
\begin{align*}
    y &\gets v - c \\
    c &\gets ((s + y) - s) - y\\
    s &\gets s + y\,,
\end{align*}
where $s$ is the current value of the sum, $c$ is the compensation term, and $v$ is the  number to be added.

At first glance, this might seem counterintuitive, as by requiring the compensation term $c$, the memory benefit of
the smaller representation for $s$ is negated. However, if $s$ were kept in high precision, that would require making
an additional low-precision copy for fast matrix multiplications, which is unnecessary with Kahan summation.

\begin{figure}[tbph]
\begin{tikzpicture}[font=\scriptsize]
    \node[anchor=south west,inner sep=0] at (0,0) {\includegraphics[width=\linewidth]{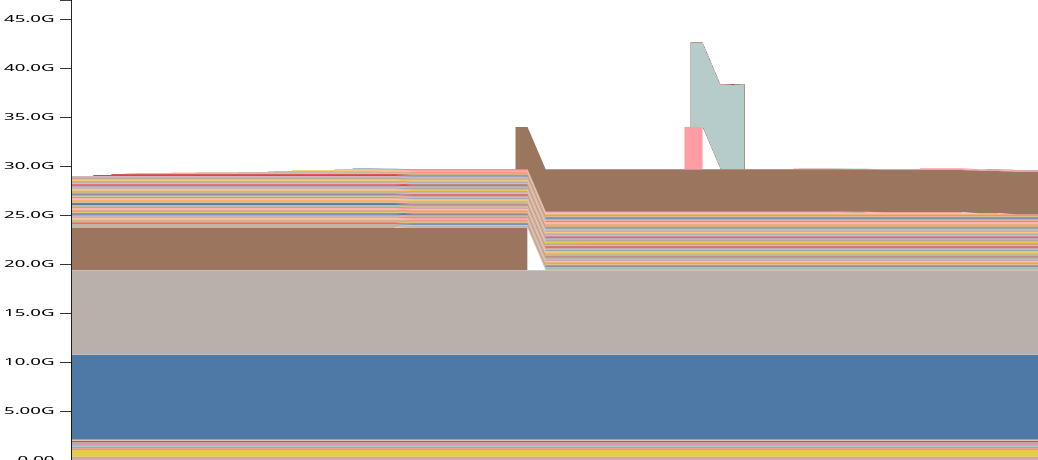}};
    \node[inner sep=0] at (5,0.5) {Classifier weights};
    \node[inner sep=0] at (5,1.2) {Momentum buffer};
    \node[inner sep=0] at (2.5,1.67) {FP16 weights $(t-1)$};
    \node[inner sep=0] at (6,2.17) {FP16 weights $(t)$};
    \node[inner sep=0] at (6,1.75) {Bert Activations};
    \node[inner sep=2] (gfp16) at (4.8,2.9) {Grad FP16};
    \node[inner sep=0] (gfp32) at (6.8,2.6) {Grad FP32};
    %\node[inner sep=0] (sgdb) at (6.8,3.5) {SGD Buffer};

    \draw[thick,->] (gfp16.south) -> (5.55,2.45);
    \draw[thick,->] (gfp32.west) -> (5.65,2.8);
    %\draw[thick,->] (sgdb.west) -> (5.8,3.3);
\end{tikzpicture}
\caption{Memory trace of \texttt{Renee}~\citep{jain2023renee} at 3 million labels \& batch size 128, recorded with Pytorch profiler}
\label{fig:memory_trace_for_renee}
\end{figure}
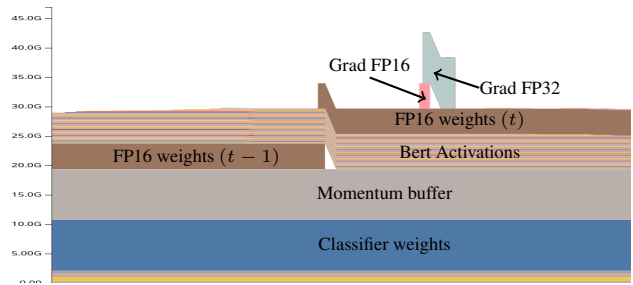

\paragraph{Shortcomings of mixed-precision training in Renee}
Mixed precision training (MPT) aims at improving throughput by ensuring that the most
compute-intensive operations, matrix multiplications (matmul), are performed in lower-precision
representations that enjoy accelerated hardware implementations. This means that
weights are kept at their full precision, and a second, ephemeral low-precision
copy of the weights is created for the purpose of matmul. In contrast, activations
are consistently kept in lower precision, and the \texttt{Float16} gradients are cast into \texttt{Float32} during weights update.

Consequently, mixed precision training greatly benefits models with moderate parameter sizes across multiple layers (e.g., layers in the BERT-base encoder),
and leads to a significant reduction in memory requirements if they are dominated by activation memory. However, its application to a single layer with extremely large parameter sizes significantly increases the peak memory consumption because it creates a second (albeit smaller)
copy of the huge classification weights.
Specifically, the memory trace for \texttt{Renee} presented in \cref{fig:memory_trace_for_renee} shows that the low-precision copies of the classifier weights
persists for the entire step.\footnote{This particular problem could easily be fixed by \texttt{del}'ing them after the calculating of the classifier layer's gradient.}
Additionally, we can see that the gradient is first calculated in 16-bit precision, but then upcast to 32 bit.\footnote{\href{https://github.com/microsoft/renee/blob/7c7c9910422c6ff092183c79cbba11cd8b9d4557/dl_base.py\#L693}{Renee official code}}

As a consequence, \texttt{Renee}'s memory consumption is considerably higher than one might na\"ively suspect, especially at its peak, with a total of about \SI{40}{\gibi\byte} at 3 million labels. Switching to pure \texttt{BFloat16} training and fusing the SGD update so it does not need an additional buffer would immediately reduce that
by at least $3\times\SI{8}{\gibi\byte}$. In the next section, we will present such a change, together with further memory optimizations, before finally reducing parameters to \texttt{FP8} for even more memory savings.

\begin{figure*}
\centering     %%% not \center
\subfigure[Precision@1 (Without SR\ With SR)]{\label{fig:sr_performance}\includegraphics[width=80mm]{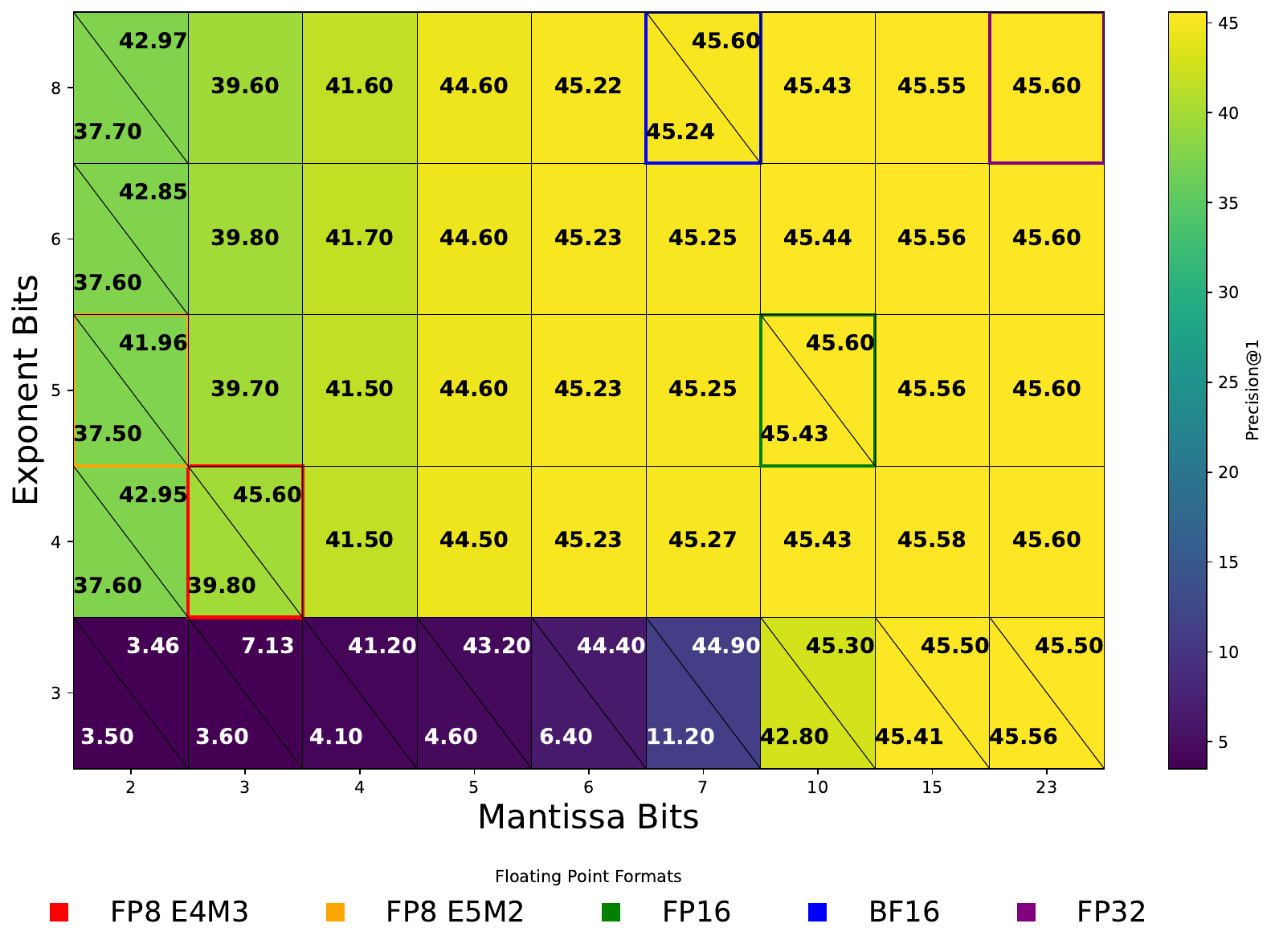}}
\subfigure[The histogram of the classifier gradient]{\label{fig:gradient}\includegraphics[width=80mm]{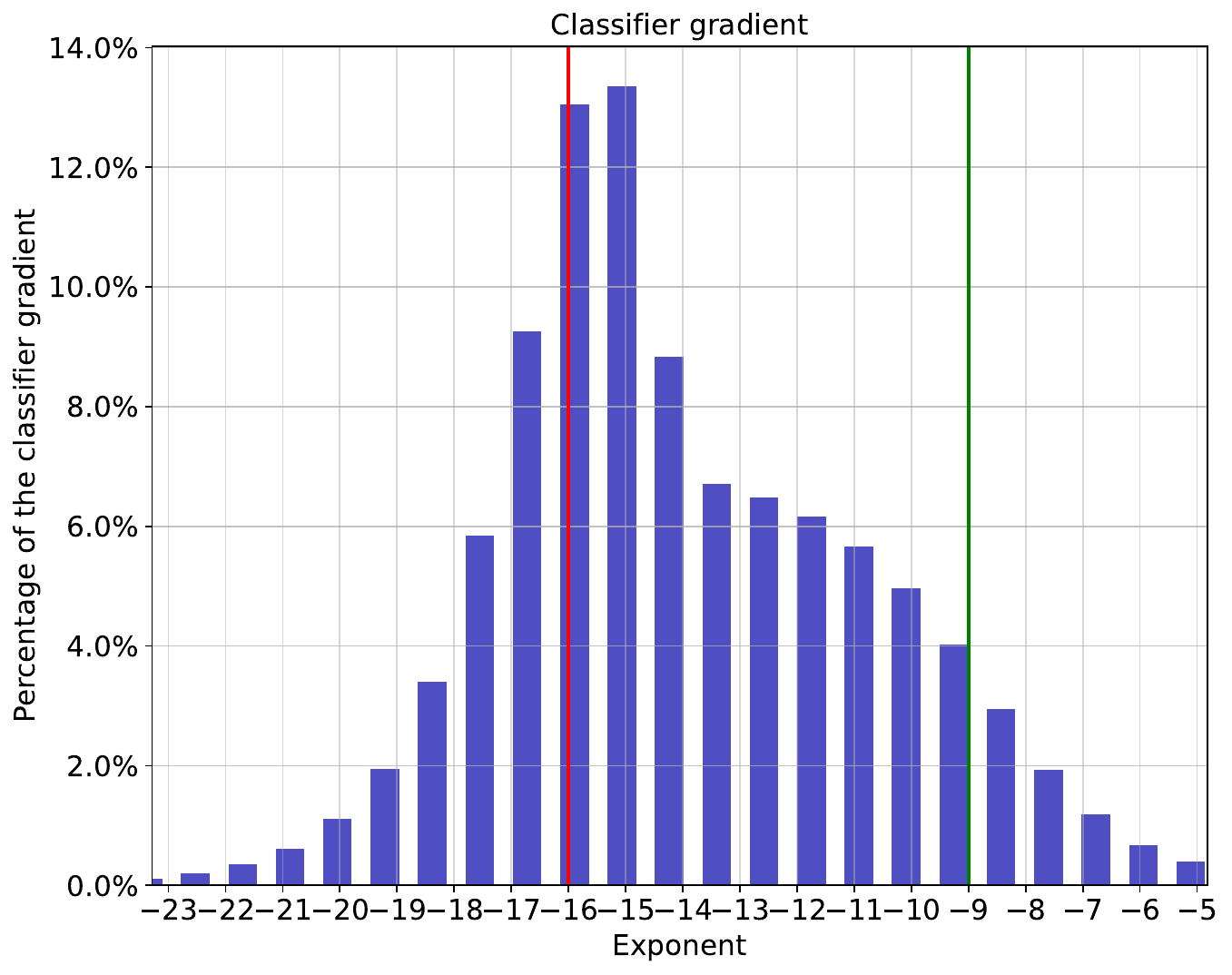}}
\caption{Figure \ref{fig:sr_performance}: Precision@1 performance at different exponent and mantissa bit patterns for the classifier weights. The numbers above diagonal is the performance when stochastic rounding is applied. Figure \ref{fig:gradient}: histogram of classifier gradients. Around 20\% of gradients become zero in \texttt{Float8 E5M2}([-16, 15]), while nearly 90\% of gradients drop to zero in \texttt{Float8 E4M3}([-9, 8]). 
% For XMC training, small updates are essential, so higher precision for the gradient is still necessary. 
The gradients are sampled from the training of LF-AmazonTitles-131K.}
\end{figure*}

\section {ELMO : Low-precision and Peak Memory Optimization in XMC}

\subsection{Pure 16-Bit Training} 
The most obvious inefficiencies visible in \cref{fig:memory_trace_for_renee} are
the multiple copies of weights and gradients in different precisions.
Thus, our first step towards memory-efficient training is the implementation of pure 16-bit training. 

This is a bit more involved than just changing the datatype of the weights tensor, as two 
problems inherent in lower bit-width representations need to be overcome. \texttt{FP16}
numbers, as used in \texttt{Renee}'s mixed precision training, have a much lower
range of representable values than the single precision baseline. Typically, this
manifests in underflows during the backward pass, leading to gradients being rounded
to zero, and is mitigated by loss-scaling techniques that shift the overall range
of gradients into the representable interval~\citep{micikevicius2018mixed}.

Gradient computation for the classifier input (i.e. classifier logit gradient $\times$ classifier weights), however, involves matrix multiplications over a large inner dimension, corresponding to the number of labels. 
This large accumulation over labels often results in \emph{overflow} outside the \texttt{FP16} range. At the same time,
gradients of the classifier weights themselves do not grow in magnitude as the label space increases. 
This contrasting nature (overflow \& underflow) of gradient behavior makes \texttt{Renee}’s training unstable and sensitive to both encoder and label space sizes despite using mixed precision with loss scaling (see Section 5 in Renee). 

One possible solution is to adopt separate scaling factors but this requires sweeping over scale combinations \cite{blake2023unit}.
Even simpler, though, overflow and underflow can be easily addressed by switching from \texttt{FP16} to \texttt{BF16},
sacrificing some precision for an extended range of values that matches \texttt{FP32}. This removes the additional
high-precision copy of the classifier's gradients.

The reduced precision of \texttt{BF16} comes at a price. If one just replaces
the 32-bit versions of weights in the optimizer with corresponding \texttt{BF16} weights, training
can no longer progress successfully. This is because round-to-nearest rounding in the optimizer update
can end up canceling the update step, if it is less than half the distance to the next representable number.

There are two common ways for dealing with this problem, and we employ both.
For the classifier weights, where memory efficiency is paramount, we use stochastic
rounding~\citep{forsythe1950round,croci2022stochastic, zamirai2020revisiting} as defined in \eqref{eq:sr}. This reduces the amount of memory required for
classifier weights to one third of the \texttt{Renee} setting and halves the optimizer state.%and eliminating the low-precision copy for the forward and backward pass.
 The weights of the \texttt{Bert} encoder, in contrast, only make up a tiny fraction of the total memory consumption.
Therefore, we opt to use \emph{Kahan summation}~\citep{kahan} for the encoder's parameters. %Both optimizer steps are implemented as a fused kernel for AdamW with Kahan summation, our own implementation for SGD with stochastic rounding, so that intermediate result is just kept in the GPUs register and never materialized in DRAM.

\subsection{ Architectural improvements }
After fixing the most egregious inefficiencies stemming from standard mixed-precision training, we now turn to further
architectural enhancements to reduce the (peak) memory consumption of the training algorithm.

A noticeable chunk of the remaining memory is taken up by the momentum buffer for the classifier weights. Our experiments
showed that momentum is not required so we remove it entirely, switching to pure large-learning-rate SGD for the classifier layer.

Finally, there remains the spike in memory consumption due to the classification layer's gradient. There are several ways
this could be addressed. A first option would be to change the order of operations and calculate this gradient only at the
end of the backward pass, when activation memory for the \texttt{Bert} encoder has already been freed. While this does not
actually reduce the memory requirements for this operation, it is moved to a point in time with less memory pressure, thus
reducing peak memory consumption.

Reorganizing computation flow helps alleviate peak memory consumption; however, as label size increases, memory allocated for classifier logits and gradients becomes the primary bottleneck. %While %recent memory-optimized LLM training strategies utilize chunking for attention, cross entropy and  loss+gradient computation, 
We employ chunking \cite{rabe2021self,hsu2024liger} for classifier parameters: We first execute the encoder's forward pass, then divide the labels into $k$ equal-sized chunks, processing the classifier's forward pass, backward pass, and optimization step sequentially for each chunk. The encoder's backward pass and update occur after all classifier operations are complete. This reduces transient memory requirements for classifier operations by a factor of $k$. We use between 3 and 8 chunks for XMC datasets, observing no impact on training latency (see Appendix \ref{chunking_ablation}).% as each chunk computation runs asynchronously between the CPU and GPU.

\begin{figure*}[ht]
    \includegraphics[width=0.99\linewidth]{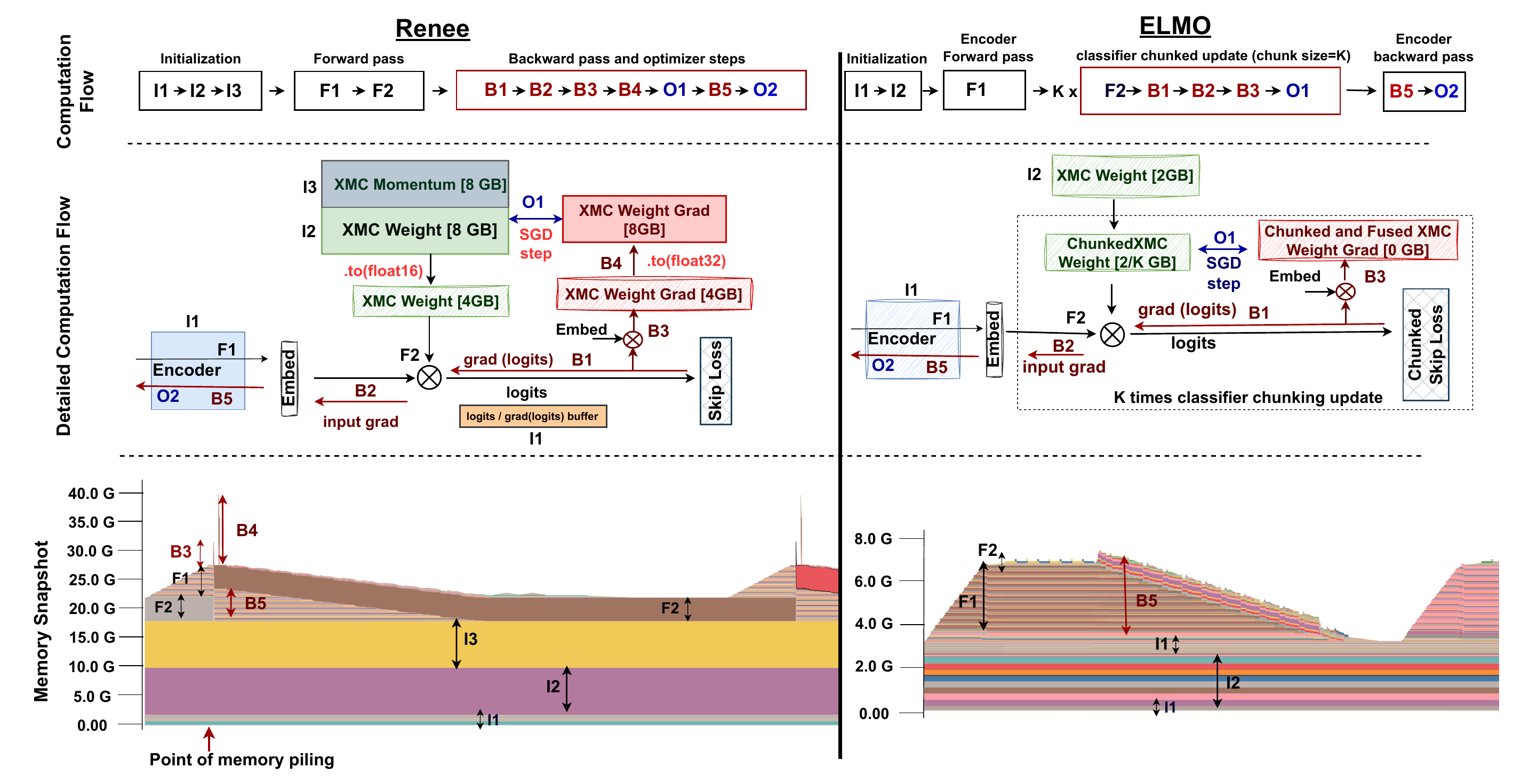}
    \caption{GPU memory comparison between Renee and the proposed approach (ELMO) at various instances during one round of forward and backward pass. Note the difference in the scale of Y-axis in the two cases. The graphic was created using Pytorch memory viz utility.}
    \label{fig:pmo}
\end{figure*}

\subsection{8-Bit Training}
For further memory savings, we turn towards reducing the number of bits allocated to each parameter further.
To that end, we first investigate how much precision and dynamic range are required for the training process,
by simulating floating-point numbers with a specific number of mantissa and exponent bits.

Figure \ref{fig:sr_performance}, which illustrates the performance for LF-AmazonTitles-131K dataset in terms of precision@$k$,
shows that 3 exponent bits provide sufficient range to represent classifier weights, whereas 2 bits are not enough. 
In the mantissa, we see significant degradation starting to set in as it shrinks below 6 bits, but this can be counteracted by stochastic rounding, which recovers the original performance. This leads us to choose the \texttt{E4M3}, which is directly supported in \texttt{Hopper}, \texttt{Ada}, and \texttt{Blackwell} GPUs, for representing classifier weights; notably, we do not introduce any additional tensor scaling~\citep{micikevicius2022fp8}, the native dynamic range of \texttt{E4M3} is sufficient, c.f. also \cref{fig:weights,fig:clf_input}.

In contrast, \cref{fig:gradient} indicates that exponents of around 20\% of the gradients exceed even the representable range of \texttt{FP8} E5M2, 
necessitating the use of \texttt{BF16}. To address the different precision requirements for weights (in \texttt{FP8}) and gradients (\texttt{BF16}), we cast classifier inputs from \texttt{BF16} to \texttt{FP8 E4M3} when computing logits, as \texttt{FP8} sufficiently covers the inputs range (Figure \ref{fig:clf_input}). We then perform a matrix multiplication between \texttt{FP8 E4M3} inputs and \texttt{FP8 E4M3} weights, but obtain logits in \texttt{BF16} for higher-precision gradient computation. For input gradient computation, which involves matrix multiplication between \texttt{FP8} weights and \texttt{BF16} logits, we developed a Triton \cite{tillet2019triton} kernel to manage this, avoiding additional HBM memory usage due to data type differences.

To reduce the memory overhead caused by the gradient, ELMO fuses gradient computation and SGD updates into a single Triton \cite{tillet2019triton} kernel, optimized for execution entirely in SRAM. Within this kernel, classifier weights, logits, and inputs are loaded from GPU memory into SRAM. Classifier weight gradients are computed via a matmul operation between the logits and inputs. The classifier weights are then updated directly in SRAM using SGD with stochastic rounding before being written back to GPU memory. This eliminates the need to store classifier gradients in GPU memory, reducing its memory footprint to nearly zero.

Further, we take on the recently developed  \texttt{torchao} \cite{torchao} framework for the encoder training, which leverages \texttt{FP8} training for transformers, reducing the encoder's activation memory requirements compared to \texttt{BF16} training. By combining \texttt{FP8} training for both the encoder and classifier, end-to-end training of \texttt{FP8} XMC models becomes feasible.

\begin{table*}[]
  \caption{Statistics of XMC Datasets with and without Label Features. This table presents a comparison across various datasets, detailing the total number of training instances ($N$), unique labels ($L$), number of test instances ($N'$), average label count per instance ($\overline{L}$), and average data points per label ($\hat{L}$).}
  \label{tab:dataset_stats}
  \centering
  \begin{small}
  \resizebox{\textwidth}{!}{%
  \begin{tabular}{
    @{}l@{\hskip 40pt}
    S[table-format=7.0,group-separator={,}]@{\hskip 50pt}
    S[table-format=6.0,group-separator={,}]@{\hskip 40pt}
    S[table-format=7.0,group-separator={,}]@{\hskip 30pt}
    S[table-format=2.2]@{\hskip 30pt}
    S[table-format=3.2]@{}
  }
    \toprule
    \textbf{Dataset} & {$N$} & {$L$} & {\textbf{$N'$}} & {\textbf{$\overline{L}$}} & {\textbf{$\hat{L}$}} \\
    \midrule
    \multicolumn{6}{c}{\textbf{Datasets without Label Features}} \\
    \midrule
    Wiki-500K           & 1779881 & 501070 & 769421  & 4.75  & 16.86 \\
    AmazonTitles-670K   & 485176  & 670091 & 150875  & 5.39  & 5.11 \\
    Amazon-670K         & 490449  & 670091 & 153025  & 5.45  & 3.99 \\
    Amazon-3M           & 1717899 & 2812281& 742507  & 36.17 & 31.64 \\
    \midrule
    \multicolumn{6}{c}{\textbf{Datasets with Label Features}} \\
    \midrule
    % LF-Amazon-131K   & 294805  & 131073 & 134835  & 5.15  & 2.29 \\
    LF-AmazonTitles-131K   & 294805  & 131073 & 134835  & 5.15  & 2.29 \\
    LF-WikiSeeAlso-320K    & 693082 & 312330 & 177515 & 4.67  & 2.11 \\
    LF-AmazonTitles-1.3M   & 2248619 & 1305265 & 970237 & 22.2& 38.24 \\
    LF-Paper2Keywords-8.6M   & 2020621 & 8623847 &2020621  & 9.03 & 2.12\\
    \bottomrule
  \end{tabular}
  
  }
  \end{small}
\end{table*}

\subsection{ Comparison of ELMO and Renee }
Finally, let us consider training a 3 million label model with a batch size of 128,
using \texttt{Bert-base} as an encoder with 768 embedding dimensions. How much memory 
do the different parts of the models take?
We use the above Figure~\ref{fig:pmo}, which shows a comparison of the memory snapshot for the two approaches, and corresponding order of operations during initialization, forward and backward pass.

At initialization (denoted by I1, I2 etc. in Figure~\ref{fig:pmo}), Renee allocates memory for encoder parameters, its optimizer states, 
and a buffer to store logit gradients for the labels in last layer. For this example setting, parameters and momentum for the classifier layer amount to \SI{8}{\gibi\byte} \footnote{ $ 8 \approx \frac{768 \times 2,812,281 \times 4} {1024 \times 1024 \times 1024} = \frac{\text{embed\_dim} \times \text{num\_labels} \times \text{num\_bytes\_per\_fp32}}{1024 \times 1024 \times 1024}$} each, the logit buffer consumes \SI{687}{\mebi\byte}. ELMO gets rid of the momentum buffer altogether,
and allocates weights in either 16-bit (\SI{4}{\gibi\byte}) or 8-bit (\SI{2}{\gibi\byte}). Due to chunking, the size of the logits' gradients gets divided by the number of chunks, in this case 8, but stays in 16-bit representation for both \texttt{BF16} and \texttt{FP8} training, leading to \SI{86}{\mebi\byte}.
Additionally, the \texttt{Bert} model and its optimizer states are allocated. These are the same size for both ELMO and Renee, amounting to $\approx\SI{1.2}{\gibi\byte}$.
In total, ELMO allocates \SI{3.2}{\gibi\byte} (\SI{5.2}{\gibi\byte} for \texttt{BF16}) at initialization, 70-80\% reduced compared to Renee's \SI{17.9}{\gibi\byte}.

During the forward propagation steps (denoted as F1, F2 etc. in Figure \ref{fig:pmo}), activation memory accumulates for the \texttt{Bert} transformer, \SI{4.6}{\gibi\byte} in \texttt{BF16} and \SI{3}{\gibi\byte} in \texttt{FP8} mixed precision. For the classifier layer, Renee also needs to create the \texttt{FP16} copy of its weights, an additional \SI{4}{\gibi\byte}. 
During the backward pass of Renee (B1, B2 etc.), gradients are allocated (\SI{4}{\gibi\byte}) and converted to \texttt{FP32} \SI{8}{\gibi\byte}, with all these allocations stacking up to a peak memory consumption of \SI{39.7}{\gibi\byte}.
Much more efficiently, ELMO does not need to make a copy of the classifier weights, nor does it materialize classifier gradients. However,
the FP8 encoder uses additional buffers of \SI{0.5}{\gibi\byte}, bringing the peak memory consumption to \SI{6.6} {\gibi\byte}.

\section{Contributed Dataset}
\textbf{Motivation.} Performance evaluation for XMC algorithms traditionally relies on public benchmarks \cite{Bhatia16}, where the largest dataset contains 3 million labels. Consequently, many recent large-scale XMC experiments have employed proprietary datasets~\citep{mehta2024astraaccuratescalableannsbased,jain2023renee,dahiya2021siamesexml}. A more expansive public dataset would facilitate the exploration and comparison of modern, efficient training strategies, while also revealing bottlenecks that become significant at larger label sizes.

\textbf{LF-Paper2Keywords-8.6M.}  We curated LF-Paper2Keywords-8.6M using the DBLP-Citation-network V14 dataset \cite{Tang:08KDD, Tang:10TKDD, Tang:11ML, Tang:12TKDE, Tang:07ICDM, sinha2015overview}. This dataset comprises the titles and abstracts of research papers sourced from DBLP, ACM, and MAG (Microsoft Academic Graph), with each paper's keywords serving as labels. The dataset includes 8.6 million labels and can support tasks such as automated keyword suggestion and paper recommendation for research articles. Complete dataset details are provided in Table \ref{tab:dataset_stats}.

\begin{table*}[h]
\caption{Comparison of the precision performance of our proposed ELMO method with state-of-the-art XMC methods on the Wiki-500K, AmazonTitles-670K, Amazon-670K, and Amazon-3M datasets, as well as the label feature datasets LF-WikiSeeAlso-320K and LF-AmazonTitles-1.3M. Bold indicates the best results, and underline indicates the second-best.  \(M_\mathrm{tr}\) denotes peak training memory and \textsc{METHOD-N} denotes a method with $N$ ensembles.}
\label{tab:main-nonlf-table}
\begin{center}
\begin{small}
\begin{tabular}{@{}l|c|ccccc|ccccc@{}}
\toprule
Method & Encoder  &  P@1 & P@3 & P@5 &   \makecell{\(M_\mathrm{tr}\)\\\((\mathrm{GiB})\)} & \makecell{Epoch Time \\ (mm:ss)} &  P@1 & P@3 & P@5 &  \makecell{\(M_\mathrm{tr}\)\\\((GiB)\)} & \makecell{Epoch Time \\ (mm:ss)} \\
\midrule \midrule
&  &  \multicolumn{5}{c|}{Wiki-500K} &  \multicolumn{5}{c}{AmazonTitles-670K} \\
\midrule

\textsc{LightXML} &  BERT-Base & 76.19 & 57.22 & 44.12 & 15.72 & 147:27 & 41.7 & 37.3 & 34.2 & 13.99 & 19:02 \\

\textsc{CascadeXML} & BERT-Base & 77.0 & 58.3 & 45.1 & 18.8 & 50:00 &  42.1 & 37.5 & 34.1 & 22.3 & 11:32 \\
\midrule

\textsc{LightXML-3} & BERT-Base & 77.78 & 58.85 & 45.57 & 15.72 & 3$\times$147:27 &  43.1 & 38.7 & 35.5 & 13.99 & 3$\times$19:02  \\
\textsc{CascadeXML-3} & BERT-Base & 78.39 & 59.86 & 46.49 & 18.8 & 3$\times$50:00 & 43.5 & 39.0 & 35.6 & 22.3 & 3$\times$11:32 \\
\midrule

\textsc{Renee} & BERT-Base & \textbf{78.69} & \underline{60.03} & \underline{46.46} & 12.69 & 18:37 & 43.78 & 39.17 & 35.91 & 12.46 & 2:04 \\
\textsc{ELMO (\texttt{BF16})} &  BERT-Base & \underline{78.61} & \textbf{60.04} & \textbf{46.59} & \underline{7.21} & \underline{17:01} &  \underline{44.3} & \underline{39.7} & \textbf{36.4} & \underline{5.12} & \underline{1:47} \\
\textsc{ELMO (\texttt{FP8})} & BERT-Base & 78.39& 59.64& 46.12& \textbf{5.01} & \textbf{11:28} &  \textbf{44.39} & \textbf{39.75} & \underline{36.43} & \textbf{3.37} & \textbf{1:18} \\
\midrule

% \midrule \midrule
& & \multicolumn{5}{c|}{Amazon-670K} &  \multicolumn{5}{c}{Amazon-3M} \\
\midrule
\textsc{LightXML} &  BERT-Base & 47.3 & 42.2 & 38.5 & 11.5 & 53:30 & - & - & - & OOM & - \\
\textsc{CascadeXML} & BERT-Base & 48.5 & 43.7 & 40.0 & 18.3 & 16:46 &  51.3 & 49.0  & 46.9 & 87.0 & 90:00  \\
\midrule
\textsc{LightXML-3} & BERT-Base & 49.1 & 44.17  & 40.25 & 11.5 & 3$\times$53:30 & - & -  & - & OOM & -  \\
\textsc{CascadeXML-3} & BERT-Base & 50.22 & \underline{45.20} & \textbf{41.45} & 18.3 & 3$\times $16:46 & 53.10 & 50.64 & 48.49 & 87.0 & 3$\times $90:00 \\
\midrule
\textsc{Renee} & BERT-Base & \underline{50.6}  & 45.16 & 41.13 & 11.91 & 7:14 & 52.6 & 49.7 & 47.43 & 39.7 & 29:58 \\

\textsc{ELMO (\texttt{BF16})} &  BERT-Base & \textbf{50.7} & \textbf{45.27} & \underline{41.29} & \underline{5.29} %4.89
& \underline{6:00} & \textbf{53.4} & \textbf{50.9} & \textbf{48.8} & \underline{10.39} & \underline{25:15}  \\
\textsc{ELMO (\texttt{FP8})} &  BERT-Base & 50.34 & 44.91 & 40.97 & \textbf{3.3} & \textbf{4:05} & 52.73 \ & 50.38 & 48.27 & \textbf{6.6} & \textbf{18:02}   \\
\midrule

&  & \multicolumn{5}{c|}{LF-WikiSeeAlso-320K} &  \multicolumn{5}{c}{LF-AmazonTitles-1.3M} \\
\midrule

\textsc{LightXML} & Distil-BERT &  34.5 & 22.31 & 16.83 & 24.46 & 61:09 & - & - & - & OOM & - \\
\textsc{NGAME} & Distil-BERT & 45.72 & 29.61 & 22.06  & 16.63 & 57:19 & 54.99 & 48.09 & 43.11  & 11.03 & 40:09 \\
\textsc{DEXML} & Distil-BERT &  46.78  &  30.32 &  22.59 & 38.6 & 242:51 & \textbf{58.4} & - & \textbf{45.46} & 75.53 & 1054:21 \\
\midrule
\textsc{Renee} & Distil-BERT &  47.86 & 31.91 & 24.05 & 13.89 & 10:50 & 56.04 & \textbf{49.91} & \underline{45.32} & 19.9 & 9:12 \\
\textsc{ELMO (\texttt{BF16})} & Distil-BERT & \underline{47.84} & \textbf{31.99} & \textbf{24.12} & \underline{6.57} & \underline{10:08} & \underline{56.14} & \underline{49.86} & 45.25 &  \underline{6.61} & 9:10 \\
\textsc{ELMO (\texttt{FP8})} & Distil-BERT & \textbf{47.88}  & \underline{31.92} & \underline{24.09} & \textbf{5.2} & \textbf{6:22} & 54.97 & 48.41 &  43.82 & \textbf{4.31}  & 17:44 \\

\bottomrule
\end{tabular}
\end{small}
\end{center}
\end{table*}

\section{Experiments and Discussion}
\noindent \textbf{Datasets.} We validated our approach on a broad suite of XMC datasets spanning both non–label-feature-based (Amazon-670K, Wiki-500K, Amazon-3M, AmazonTitles-670K) and label-feature-based (LF-AmazonTitles-131K, LF-WikiSeeAlso-320K, LF-AmazonTitles-1.3M, and our newly curated LF-Paper2Keywords-8.6M). All except LF-Paper2Keywords-8.6M are publicly accessible through the Extreme Classification Repository \cite{Bhatia16}. Detailed descriptions are provided in the Table \ref{tab:dataset_stats}.

\noindent \textbf{Baselines and Evaluation Metrics.} We compared our method with two categories of baselines: \emph{(i) Sampling-Based XMC}, focusing on Transformer-based methods (e.g., LightXML \cite{jiang2021lightxml}, CascadeXML \cite{kharbanda2022cascadexml}) for non-label-feature datasets and NGAME \cite{dahiya2023ngame} and DEXML \cite{gupta2024dualencoders} for label-feature datasets; \emph{(ii) End-to-End XMC}, with Renee \cite{jain2023renee} as the principal baseline. Following standard XMC practices, we evaluate all methods using top-\emph{k} metrics, specifically Precision@\emph{k} and its propensity-scored variant. A detailed overview is given in the Appendix \ref{metrics}.

\begin{table}[ht]
\caption{Precision performance comparison on LF-Paper2Keywords-8.6M dataset. Other XMC baselines do not scale to $8.6$ million label size.}
\label{tab:lf-table}
\begin{center}
\resizebox{0.49\textwidth}{!}{%
%\begin{small}
\begin{tabular}{@{}l|cccc}
\toprule
Method &  P@1 & P@3 & P@5 &   \makecell{\(M_\mathrm{tr}\)\((\mathrm{GiB})\)} \\
\midrule \midrule
\textsc{Float32} &  \underline{43.60} & \underline{32.13} & \underline{26.02} & 58.44   \\
\textsc{Renee} & 17.65  & 11.78 & 9.23& 105.64   \\
\textsc{ELMO(\texttt{BF16})} & \textbf{45.4}  & \textbf{33.58}  &  \textbf{27.18} &  \underline{18.8} \\
\textsc{ELMO(\texttt{FP8})} & 43.4 & 31.59& 25.38 & \textbf{9.02}  \\
\bottomrule
\end{tabular}
%\end{small}
}
\end{center}
\end{table}

\noindent \textbf{Implementation Details.} Low-precision training with chunking is implemented using the PyTorch framework \cite{paszke2017automatic}. For the encoder, we used AdamW \cite{loshchilovdecoupled} optimizer with Kahan summation provided by the \texttt{optimi} library\footnote{\href{https://github.com/warner-benjamin/optimi}{\texttt{https://github.com/warner-benjamin/optimi}}}. The in-place SGD optimizer with stochastic rounding for the classifier is implemented via custom Triton and CUDA kernels. We also employ Triton kernels for the \texttt{FP8} classifier and gradient fusion. %\footnote{\href{https://github.com/xmc-aalto/elmo}{ELMO official code}}. 
All \texttt{BF16} experiments are conducted on an A100 GPU, while \texttt{FP8} experiments are run on an H100 (Table \ref{tab:main-nonlf-table}, \ref{tab:lf-table} and \ref{tab:131k})  and RTX 4060Ti (Table \ref{tab:rxt4060}) GPU. Further details on implementation and hyperparameters are provided in the Appendix \ref{hyperandimplementations}.

\noindent \textbf{Empirical Performance.} 
Table \ref{tab:main-nonlf-table} reports Precision@\emph{k} results for ELMO on non–label-feature datasets, benchmarked against state-of-the-art XMC baselines. Notably, ELMO significantly reduces peak memory usage---its FP8 variant achieves a 6 times reduction relative to Renee \cite{jain2023renee} (the current end-to-end optimized approach) and a 13 times reduction compared to sampling-based methods. Despite these efficiency gains, ELMO shows competitive or improved Precision@\emph{k} performance. For label-feature datasets, we apply the standard augmentation \cite{kharbanda2024gandalf,jain2023renee} strategy summarized in Table \ref{tab:main-nonlf-table}  (LF-WikiSeeAlso-320K, LF-AmazonTitles-1.3M) and Table \ref{tab:131k}  (LF-AmazonTitles-131K). Our results demonstrate that ELMO maintains robust performance in both FP8 and BF16 modes, underscoring the versatility of our approach.

Finally, on our newly introduced LF-Paper2Keywords-8.6M dataset as shown in Table \ref{tab:lf-table}, ELMO provides substantial memory savings, requiring only  \SI{18.8}{\gibi\byte} (\texttt{BF16}) or \SI{9.02}{\gibi\byte} (\texttt{FP8}), compared to \SI{105}{\gibi\byte}  for Renee. Notably, \texttt{BF16} ELMO even outperforms the \texttt{Float32} baseline, likely benefiting from the regularization effects of stochastic rounding \cite{ozkara2025stochastic}. \texttt{FP8} also delivers performance close to that of \texttt{Float32}. %Renee underperforms, likely due to gradient clipping issues arising from its \texttt{FP16} data types (mixed precision discards certain updates). % 
Renee underperforms, likely due to gradient overflow in the classifier input caused by its use of \texttt{FP16} data types.

\label{fig:peak}
\begin{figure}[ht]
    \includegraphics[width=\linewidth]{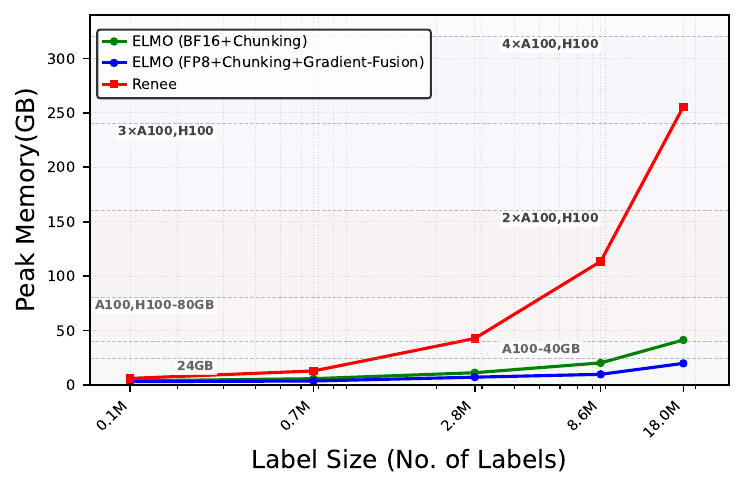}
    \caption{Comparison of peak GPU memory usage across varying label sizes for ELMO and Renee \cite{jain2023renee}.}
    \label{fig:label_size_vs_mem}
\end{figure}

\noindent \textbf{Label Size vs Peak GPU Memory.} Figure \ref{fig:label_size_vs_mem} plots peak GPU memory usage as label sizes grow from 131K (LF-AmazonTitles-131K) to 18 million. While the largest public dataset has 3 million labels, we introduce one with 8.6 million. Beyond 8.6 million, random labels are appended to measure peak usage. \texttt{BF16} and \texttt{FP8} significantly reduce GPU memory compared to Renee \cite{jain2023renee}; for instance, at 3 million labels, ELMO (\texttt{FP8}) lowers memory by 6 times, increasing to 11 times at 8.6 million and 13 times at 18 million.

\noindent \textbf{\texttt{BFloat16} vs. \texttt{Float8} Encoders.} For our encoder, we can also use \texttt{BF16} instead of the \texttt{FP8} encoder from \texttt{torchao} \cite{torchao} , while the XMC classifier continues to use \texttt{FP8}. We compare the performance with different precision settings in Table \ref{diff_encoder}, which shows similar precision but longer epoch time for FP8 due to the overhead in the FP8-BF16 mixed precision recipe.

\begin{table}[h]
\caption{Comparison of precision performance between the \texttt{BFloat16} and \texttt{Float8} encoders with the classifier fixed at \texttt{Float8}.}
\label{diff_encoder}
\begin{center}
\begin{small}
\begin{tabular}{@{}l|ccccc}
\toprule
Encoder &  P@1 & P@3 & P@5 &   \makecell{\(M_\mathrm{tr}\)\\\((\mathrm{GB})\)} & \makecell{Epoch Time \\ (mm:ss)}  \\
\midrule \midrule
&  \multicolumn{5}{c}{LF-AmazonTitles-1.3M} \\
\midrule
\textsc{\texttt{BF16}} & \textbf{55.08} & \textbf{48.47} & \textbf{43.87} &  5.50 & \textbf{17:26} \\
\textsc{\texttt{FP8}}  & 54.97 & 48.41 &  43.82 & \textbf{4.63}  & 17:44 \\
\midrule
&  \multicolumn{5}{c}{Amazon-3M} \\
\midrule
\textsc{\texttt{BF16}} & 52.60 & 50.23 & 48.11 & 8.51 & \textbf{15:56} \\
\textsc{\texttt{FP8}} & \textbf{52.73} & \textbf{50.38} & \textbf{48.27} & \textbf{7.16} & 18:02 \\
\bottomrule
\end{tabular}
\end{small}
\end{center}
\end{table}

\section{Conclusion} We present a low-precision training framework with \texttt{BFloat16} and \texttt{Float8} for XMC models, moving beyond the conventional practice of relaxing the classification layer’s precision. In contrast to \texttt{FP16-FP32} MPT which can be memory-inefficient and unstable, our method is robust and, with gradient fusion and chunking, reduces memory by 4x–6x for a 3-million-label dataset. 
Notably, these efficiency gains do not compromise performance; our approach competes with existing XMC baselines on most public datasets. 
Furthermore, we introduce a new XMC dataset LF-Paper2Keywords-8.6M with 8.6 million labels, which, upon its release, will be the largest publicly available XMC dataset. We anticipate that this new dataset will spur further innovation in extreme classification. While our proposed training recipe does not require (micro-)tensor scaling, our investigation in \autoref{fig:sr_performance} indicates that future work aiming to further reduce representation bitwidth to FP6 or FP4 datatypes will have to take such strategies into account.

\section*{Impact Statement}
We do not anticipate any negative societal impact of our work.
It is expected that the affordable training methodologies developed in this work will further enable the exploration of similar methodologies for other deep networks which are more affordable and easily accessible to a broader research community.
\section*{Acknowledgements}
We acknowledge the support of the Academy of Finland (Research Council of Finland) via grants 347707 and 348215 and the support of computational resources provided by the Aalto Science-IT project, and CSC IT Center for Science, Finland.

% Authors are \textbf{required} to include a statement of the potential 
% broader impact of their work, including its ethical aspects and future 
% societal consequences. This statement should be in an unnumbered 
% section at the end of the paper (co-located with Acknowledgements -- 
% the two may appear in either order, but both must be before References), 
% and does not count toward the paper page limit. In many cases, where 
% the ethical impacts and expected societal implications are those that 
% are well established when advancing the field of Machine Learning, 
% substantial discussion is not required, and a simple statement such 
% as the following will suffice:

% ``This paper presents work whose goal is to advance the field of 
% Machine Learning. There are many potential societal consequences 
% of our work, none which we feel must be specifically highlighted here.''

% The above statement can be used verbatim in such cases, but we 
% encourage authors to think about whether there is content which does 
% warrant further discussion, as this statement will be apparent if the 
% paper is later flagged for ethics review.


\bibliography{main_paper}
\bibliographystyle{icml2025}

%%%%%%%%%%%%%%%%%%%%%%%%%%%%%%%%%%%%%%%%%%%%%%%%%%%%%%%%%%%%%%%%%%%%%%%%%%%%%%%
%%%%%%%%%%%%%%%%%%%%%%%%%%%%%%%%%%%%%%%%%%%%%%%%%%%%%%%%%%%%%%%%%%%%%%%%%%%%%%%
% APPENDIX
%%%%%%%%%%%%%%%%%%%%%%%%%%%%%%%%%%%%%%%%%%%%%%%%%%%%%%%%%%%%%%%%%%%%%%%%%%%%%%%
%%%%%%%%%%%%%%%%%%%%%%%%%%%%%%%%%%%%%%%%%%%%%%%%%%%%%%%%%%%%%%%%%%%%%%%%%%%%%%%
\clearpage
\appendix

% \begin{table*}[h]
% \caption{Comparison of Precision performance, Peak training memory and epoch time with different baselines}
% \label{tab:main-lf-table}
% \begin{center}
% \begin{small}
% \begin{tabular}{@{}l|c|ccccc|ccccc@{}}
% \toprule
% Method & Encoder  &  P@1 & P@3 & P@5 &   \makecell{\(M_\mathrm{tr}\)\\\((\mathrm{GiB})\)} & \makecell{Epoch Time \\ (mm:ss)} &  P@1 & P@3 & P@5 &  \makecell{\(M_\mathrm{tr}\)\\\((GiB)\)} & \makecell{Epoch Time \\ (mm:ss)} \\
% \midrule \midrule

% &  & \multicolumn{5}{c|}{LF-WikiSeeAlso-320K} &  \multicolumn{5}{c}{LF-AmazonTitles-1.3M} \\
% \midrule

% \textsc{LightXML} & Distil-BERT &  34.5 & 22.31 & 16.83 & 29.97 & 143:52 & - & - & - & OOM & - \\
% \textsc{NGAME} & Distil-BERT & 45.72 & 29.61 & 22.06  & 17.86 & 57:19 & 54.99 & 48.09 & 43.11  & 11.85 & 40:09 \\
% \textsc{DEXML} & Distil-BERT &  46.78  &  30.32 &  22.59 & 41.45 & 242:51 & \textbf{58.4} & - & \textbf{45.46} & 81.1 & 1054:21 \\
% \midrule
% \textsc{Renee} & Distil-BERT &  47.86 & 31.91 & 24.05 & 13.89 & 10:50 & 56.04 & \textbf{49.91} & \underline{45.32} & 19.9 & 9:12 \\
% \textsc{ELMO (\texttt{BF16})} & Distil-BERT & \underline{47.84} & \textbf{31.99} & \textbf{24.12} & \underline{6.57} & \underline{10:08} & \underline{56.14} & \underline{49.86} & 45.25 &  \underline{6.61} & 9:10 \\
% \textsc{ELMO (\texttt{FP8})} & Distil-BERT & \textbf{47.88}  & \underline{31.92} & \underline{24.09} & \textbf{5.58} & \textbf{6:22} & 54.97 & 48.41 &  43.82 & \textbf{4.63}  & 17:44 \\

% \bottomrule
% \end{tabular}
% \end{small}
% \end{center}
% \end{table*}

\section{Baselines and Evaluation Metrics} \label{metrics}

 We compare our method with deep XMC methods with mainly transformer encoder.
 \begin{itemize}
    \item \textbf{LightXML} \cite{jiang2021lightxml}: The method employs a transformer encoder to concurrently train both the retriever and ranker, which incorporates dynamic negative sampling to enhance the model's efficacy.
    \item \textbf{CascadeXML} \cite{kharbanda2022cascadexml}:  This method separates the feature learning of distinct tasks across various layers of the Probabilistic Label Tree (PLT) and aligns them with corresponding layers of the transformer encoder. 
    \item \textbf{NGAME} \cite{dahiya2023ngame}: NGAME enhances transformer-based training for extreme classification by introducing a negative mining-aware mini-batching technique, which supports larger batch sizes and accelerates convergence by optimizing the handling of negative samples.
    \item \textbf{Renee} \cite{jain2023renee}: The Renee model employs an integrated end-to-end training approach for extreme classification, using a novel loss shortcut for memory optimization and a hybrid data-model parallel architecture to enhance training efficiency and scalability.
    \item \textbf{DEXML} \cite{gupta2024dualencoders}: The DEXML model aims to eliminate the need for an explicit classifier, instead relying solely on dual-encoder-based training with all negative labels within each batch. While the motivation to remove the classifier (often the primary bottleneck) is sound, their approach ultimately incurs higher computational and memory costs compared to methods that retain a classifier.
\end{itemize}

To evaluate the performance of our Extreme Multi-label Classification  model, which incorporates low-precision training, we use a set of metrics designed to provide a comprehensive analysis of both overall and label-specific model performance. The primary metrics we employ is Precision at $k$ (P@$k$), which assess the accuracy of the top-$k$ predictions. Additionally, we incorporate Propensity-Scored Precision at $k$ (PSP@$k$) to gauge the uniformity of the model's effectiveness across the diverse range of labels typical in XMC problems.
\paragraph{Precision at $k$ (P@$k$):} Precision at k is the fundamental metric for evaluating the top-$k$ predictions in XMC applications such as e-commerce product recommendation and document tagging:
\begin{equation}
    P@k(y, \hat{y}) = \frac{1}{k} \sum_{\ell \in \text{top}_k(\hat{y})} y_\ell
\end{equation}
where $y$ is the true label vector, $\hat{y}$ is the predicted score vector, and $ \text{top}_k(\hat{y})$ identifies the indices with the top-$k$ highest predicted scores.

\paragraph{Propensity-Scored Precision at $k$ (PSP@$k$):} Given the long-tailed label distribution in many XMC datasets, PSP@k incorporates a propensity score 
$y_l$ to weight the precision contribution of each label, thereby emphasizing the tail labels' performance:
\begin{equation}
    PSP@k(y, \hat{y}) = \frac{1}{k} \sum_{\ell \in \text{top}_k(\hat{y})} \frac{y_\ell}{p_\ell}
\end{equation}
where $p_l$ corresponds to the propensity score for the label $y_l$ \cite{jain2016extreme}.

\begin{algorithm}
\caption{\texttt{Float8} XMC classifier}
\label{alg:fp8}
\definecolor{codeblue}{rgb}{0.25,0.5,0.5}
\definecolor{codeblue2}{rgb}{0,0,1}
\lstset{
  backgroundcolor=\color{white},
  basicstyle=\fontsize{7.2pt}{7.2pt}\ttfamily\selectfont,
  columns=fullflexible,
  breaklines=true,
  captionpos=b,
  commentstyle=\fontsize{7.2pt}{7.2pt}\color{codeblue},
  keywordstyle=\fontsize{7.2pt}{7.2pt}\color{codeblue2},
    emph={matmul_8bit, X_rowwise, W_global, G_rowwise, W_int8, G_int8, X_int8},
    emphstyle={\color[RGB]{255,52,179}},
    moreemph=[2]{matmul_16bit},
    emphstyle=[2]{\color{black}},
}
\begin{lstlisting}[language=python]
class FP8Classifier(nn.Module):
    def XMC_update(self, X, labels):
        # X [b, m] inputs for the classifier
        # Rows and cols of the positive labels
        X_gradient = torch.zeros_like(X)
        rows, cols = labels[:,0], labels[:,1]
        for i in range(self.num_chunks):
            # self.W[i] [n, m], the classifier weights of one chunk
            logits = matmul_fp8(
                self.W[i], X.t().to(torch.float8_e4m3fn)
            ) # The logits are in BF16
            rows_i, cols_i = filter_chunk_i_labels(rows, cols)
            logits = logits.sigmoid_()
            logits[cols_i, rows_i] -= 1
            X_gradient += large_k_matmul(logits,
                                                self.W[i])
        
            fuse_update(self.W[i], self.lr, logits, X, X.shape[0], self.bs, self.bs, self.bs)
        return X_gradient
        
    def fuse_update(W, lr, logits, X, K, bk, bm, bn):
        # the pseudo code of the triton kernel
        grad = tl.zeros((bm, bn), dtype=FP32)
        for _  in range(0, K, bk):
            x = load_block_from_HBM(X).to(BF16)
            l = load_block_from_HBM(logits).to(BF16)
            grad += block_matmul(l, x)
        w = load_block_from_HBM(W).to(FP32)
        w = w - lr*grad
        w = stochastic_rounding_to_FP8(w)
        write_to_HBM(w, W)
\end{lstlisting}
\end{algorithm}

\begin{figure*}
\centering     %%% not \center
\subfigure[The histogram of the classifier weights]{\label{fig:weights}\includegraphics[width=75mm]{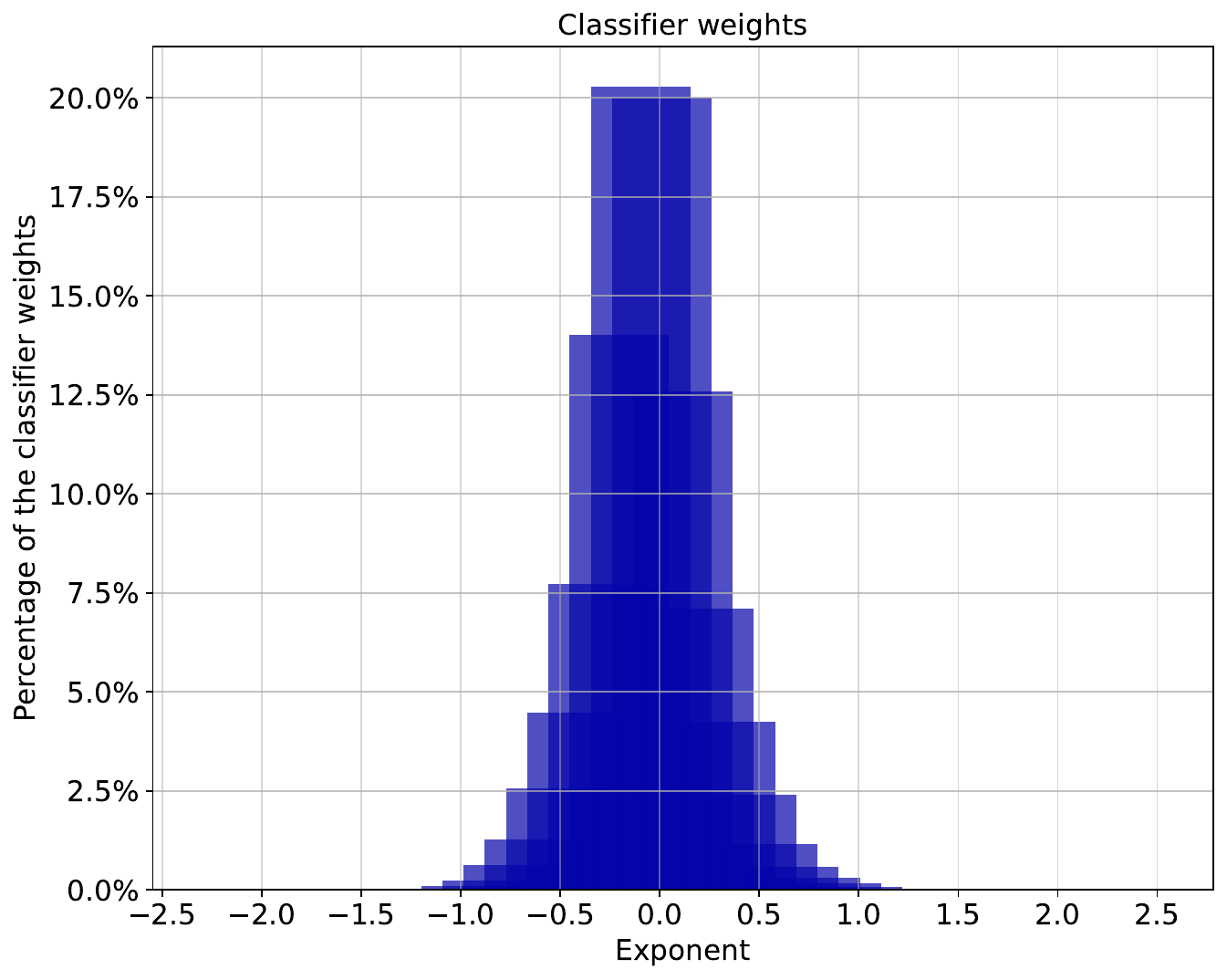}}
\subfigure[The histogram of the classifier input]{\label{fig:clf_input}\includegraphics[width=75mm]{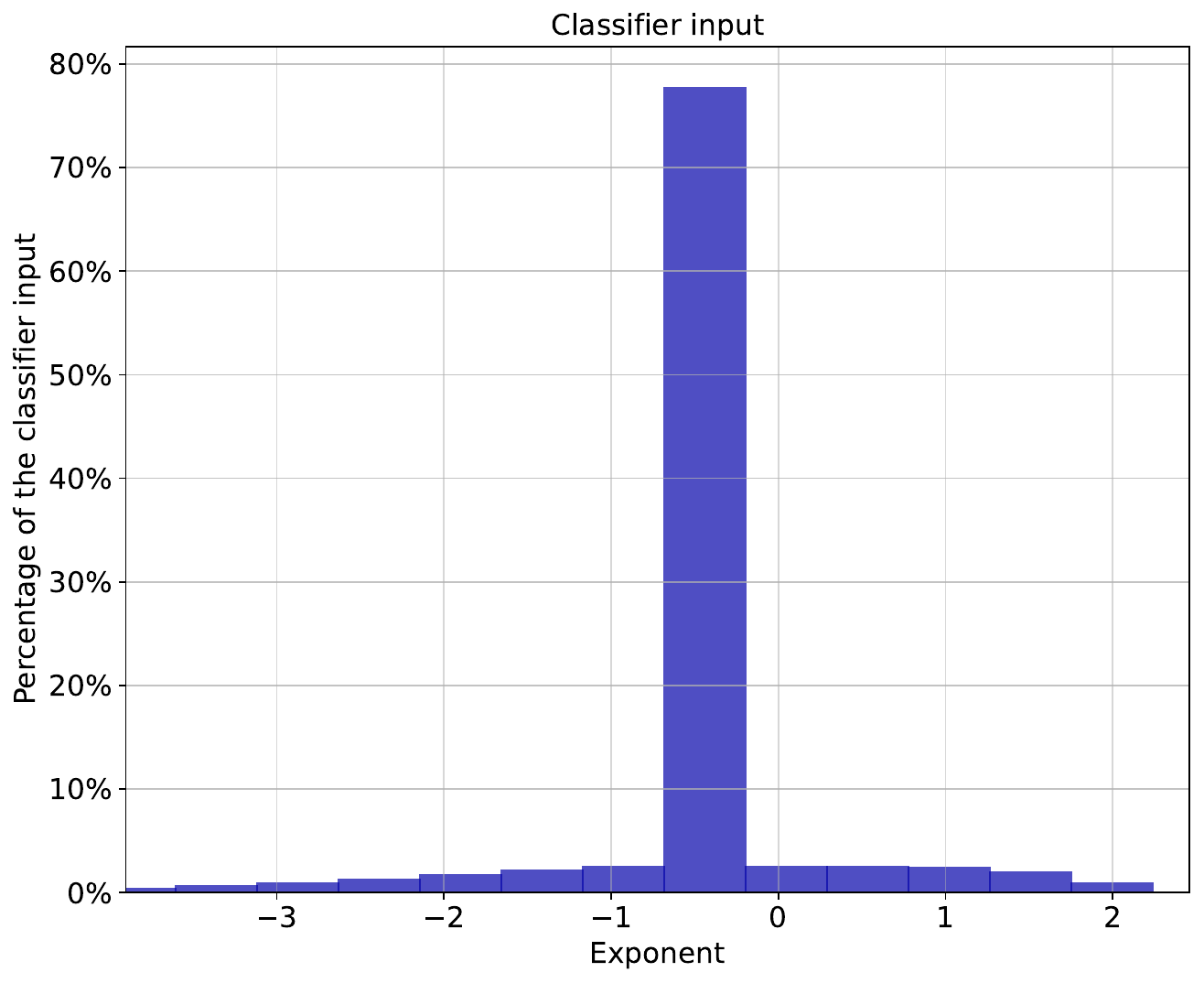}}
\caption{Figures \ref{fig:weights} and \ref{fig:clf_input} show that most weights and classifier inputs fall within the exponent range of \texttt{FP8} E4M3 ([-9, 8]), even without quantization. The weights and inputs are sampled from the training of LF-AmazonTitles-131K.}
\end{figure*}

\begin{table}[ht]
\centering
\caption{Epoch level training statistics on the RTX 4060 Ti. \(M_\mathrm{tr}\)\ denotes peak training memory. }
\label{tab:rxt4060}
\resizebox{0.48\textwidth}{!}{%
\begin{tabular}{l|cc}
\toprule
\textbf{Dataset} & \textbf{Epoch Time (mm:ss)} & \textbf{\makecell{\(M_\mathrm{tr}\)\((GB)\)}} \\
\midrule \midrule
LF-AmazonTitles-1.3M   &57:36 &  5.45  \\
\midrule
Amazon-3M   & 121:17 &  8.46\\
\midrule
LF-Paper2Keywords-8.6M  & 229:24 & 10.49 \\
\bottomrule
\end{tabular}
}
\end{table}

\begin{table}[ht]
\centering
\caption{Precision performance with Post-Hoc classifier refinement (on top of \texttt{Float8} checkpoint from ELMO ) and Kahan summation for head labels (20\% head) on the LF-AmazonTitles-1.3M. \(M_\mathrm{tr}\)\ denotes peak training memory.}
\label{tab:khan}
\resizebox{0.48\textwidth}{!}{%
\begin{tabular}{c|cccc}
\toprule
 & P@1 & p@3 & P@5 & \textbf{\makecell{\(M_\mathrm{tr}\)\((GiB)\)}} \\
\midrule \midrule
Renee   &56.04  &49.91  &45.32   &19.9  \\
\midrule
BF16 (ELMO)  &56.14 & 49.86&45.25 &6.61 \\
Float8 (ELMO)   &54.97 &48.41 &  43.82& 4.31\\
\midrule
Post-Hoc  & 55.4 & 48.87  & 44.34 & 4.31 \\
Head Kahan  & 55.6 & 49.38  & 44.88& 4.65 \\
\bottomrule
\end{tabular}
}
\end{table}

\begin{table*}[ht]
\caption{Comparison of Propensity based Precision@k for different XMC methods. The best results are denoted by bold and second best results are denoted by underline. }
\label{tab:psp-table}
\begin{center}
\begin{small}
\resizebox{\textwidth}{!}{%
\begin{tabular}{@{}l|ccc|ccc|ccc@{}}
\toprule
Method &  PSP@1 & PSP@3 & PSP@5 &  PSP@1 & PSP@3 & PSP@5 &  PSP@1 & PSP@3 & PSP@5  \\
\midrule \midrule
 &   \multicolumn{3}{c|}{Wiki-500K} &  \multicolumn{3}{c|}{AmazonTitles-670K} & \multicolumn{3}{c}{Amazon-670K} \\
\midrule

\textsc{AttentionXML} & 30.69 & 38.92 & 44 & 24.24 & 26.43 & 28.39 &  30.29 & 33.85 & 37.13  \\
\textsc{XR-Transformer} & 32.1 & 39.41 & 43.75 & - & - & - &  29.21 & 33.49 & 37.65  \\
\textsc{CascadeXML} & 31.25 & 39.35 & 43.29 & - & - & - &  30.23 & 34.93 & 38.79  \\
\midrule

\textsc{Renee} & \underline{32.9} & \underline{42.31} & \underline{46.78} & 27 & 31.1 & 34.89 & \textbf{31.45} & \textbf{36.16} & \textbf{40.15}  \\
\textsc{ELMO (\texttt{BF16})} &  \textbf{33.32} & \textbf{42.56} & \textbf{47.03} & \textbf{28.62} & \textbf{32.13} & \textbf{35.27} &  \underline{30.84} & \underline{35.69} & \underline{40.06}  \\
\textsc{ELMO (\texttt{FP8})} & 32.40 & 41.68 & 46.17 & \underline{28.24} & \underline{31.88} & \underline{35.26} &  30.57 & 35.33 & 39.67 \\
\midrule
 &   \multicolumn{3}{c|}{Amazon-3M} &  \multicolumn{3}{c|}{LF-WikiSeeAlso-320K} & \multicolumn{3}{c}{LF-AmazonTitles-1.3M} \\
\midrule
\textsc{LightXML} & - & - & - & 17.85 & 21.26 & 24.16 &  - & -  & -   \\
\textsc{XR-Transformer} & - & - & - & 25.18 & 30.13 & 33.79 &  20.06 & 24.85 & 27.79  \\
\textsc{NGAME-2} & - & - & - & \textbf{33.83} & \textbf{37.79} & \underline{41.03} & 29.18 & 33.01 & 35.36  \\

\midrule
\textsc{Renee} & 14.39 & 17.47  & 19.80 & \underline{32.02} & 37.07 & 40.9 & \underline{28.54} & \underline{33.38} & \underline{36.14}  \\

\textsc{ELMO (\texttt{BF16})} &  \underline{15.65} & \underline{19.05} & \underline{21.6} & 31.65 & \underline{37.08}
& \textbf{41.04} & \textbf{30.38} & \textbf{34.59} & \textbf{37.09}   \\
\textsc{ELMO (\texttt{FP8})} & \textbf{16.06}& \textbf{19.48} & \textbf{21.98} & 31.87& 36.98 & 40.90 & 26.72 & 31.58 & 34.46  \\
\bottomrule
\end{tabular}
}
\end{small}
\end{center}
\end{table*}

\begin{table*}[t]
\caption{Precision and propensity scored precision comparison of ELMO with state of the art XMC baselines on LF-AmazonTitles-131K dataset. The best results are denoted by bold and second best  are by underline. \(M_\mathrm{tr}\)\ denotes peak training memory.}
\label{tab:131k}
\begin{center}
\begin{small}
\resizebox{\textwidth}{!}{%
\begin{tabular}{@{}l|cccccccc@{}}
\toprule
Method &  P@1 & P@3 & P@5 & PSP@1 & PSP@3 & PSP@5   & \makecell{\(M_\mathrm{tr}\)\((\mathrm{GiB})\)} & \makecell{Epoch Time (mm:ss)}  \\
\midrule
\midrule
\textsc{LightXML} &  35.6 & 24.15 & 17.45 & 25.67  & 31.66 & 36.44 & 16.79 & 10:27   \\
\textsc{NGAME} &   44.69 & 29.89 & 21.21 & 38.81 & 44.4 & 49.43 & 11.03 & 5:15 \\
\textsc{DEXML} &   42.52 & - & 20.64 & - & - & 48.7 & 29.22 & 14:08   \\
\midrule

\textsc{Renee} &  \textbf{46.05} & \textbf{30.81} & \textbf{22.04} & \textbf{39.08} & \textbf{45.12} & \textbf{50.48} &  5.53 & 0:33 \\
\textsc{ELMO (\texttt{BF16})} &  \underline{45.6} & \underline{30.6} & \underline{21.9} & \underline{38.84} & \underline{45.02} & 50.39 & \underline{3.41} & \underline{0:31}  \\
\textsc{ELMO (\texttt{FP8})} & 45.45 & 30.53 & 21.87 & 38.75& 44.98 & \underline{50.41} & \textbf{2.75} & \textbf{0:22} \\

\bottomrule
\end{tabular}
}
\end{small}
\end{center}

\end{table*}

\section{Background of Skipping Loss Computation}
For the XMC problems in this paper, we use the binary cross-entropy loss. Following Renee \cite{jain2023renee}, we apply skipping loss computation, so the gradient for the classifier input is given by:  

\[
\text{InputGrad} = \left(\frac{1}{1 + e^{-y}} - Y\right) * W
\]

where \( y \) represents the logits, \( Y \) is the ground truth, and \( W \) denotes the classifier weights.  

Similarly, the gradient for \( W \) is:  

\[
\nabla W = \left(\frac{1}{1 + e^{-y}} - Y\right) * X
\]

where \( X \) is the input embedding for the classifier.  

We refer to \( \left(\frac{1}{1 + e^{-y}} - Y\right) \) as the "classifier logit gradient" in our paper.

\begin{table*}[t]
\centering
\caption{The hyper-parameters of the \texttt{BFloat16} and \texttt{Float8} models. Dropout is the embedding dropout and Encoder LR and XMC LR are the learning rates for the encoder and classifiers . WD denotes weight decay.}
\label{tab:hyper}
\resizebox{\textwidth}{!}{%
\begin{tabular}{l|ccccccccc}
\toprule
\textbf{Dataset} & \textbf{Encoder} & \textbf{Batch Size} & \textbf{Seq. Length} & \textbf{Dropout} &  \textbf{Encoder LR} & \textbf{XMC LR} & \textbf{Epochs} & \textbf{Warmup} & \textbf{WD (Encoder, XMC)}  \\
\midrule \midrule
 &   \multicolumn{9}{c}{\textbf{Dataset without Label Features}}  \\
 \midrule

Wiki-500K (\textsc{\texttt{BF16}}) & BERT-Base  & 128  & 128  & 0.65  & 0.00002  & 0.05  & 35  &  1000 & (0.01, 0.0001)  \\
Wiki-500K (\textsc{\texttt{FP8}}) & BERT-Base  & 128 & 128  &  0.65&  0.00002&0.15  &70  &1000  &  (0.01, 0.0001)\\
\midrule
AmazonTitles-670K (\textsc{\texttt{BF16}}) & BERT-Base  & 256  & 32  & 0.7  & 0.00005   & 0.05  & 100  & 1000 & (0.01, 0.0001) \\
AmazonTitles-670K (\textsc{\texttt{FP8}}) & BERT-Base  & 256 & 32  & 0.75 &0.00005  & 0.05 & 150& 1000 & (0.01, 0.0001)   
 \\
\midrule
Amazon-670K (\textsc{\texttt{BF16}}) & BERT-Base  & 64  & 128  & 0.75  & 0.00002  & 0.06  & 100  & 500  & (0.01, 0.0001)   \\
Amazon-670K (\textsc{\texttt{FP8}}) & BERT-Base  & 64  & 128  & 0.7 & 0.00002 & 0.05 & 150 &1000  &  (0.01, 0.0001) \\
\midrule
Amazon-3M (\textsc{\texttt{BF16}}) & BERT-Base  & 128  & 128  & 0.6  & 0.00005 & 0.05  & 90 & 10000 & (0.001, 0.001)  \\
Amazon-3M (\textsc{\texttt{FP8}}) & BERT-Base  & 128 & 128  & 0.65 & 0.00002 & 0.05 & 150 & 10000 &   (0.001, 0.001)\\
\midrule
 &   \multicolumn{9}{c}{\textbf{Dataset with Label Features}}  \\
 \midrule
LF-AmazonTitles-131K (\textsc{\texttt{BF16}}) & Distil-BERT   & 512  & 32  & 0.84  & 0.00001  & 0.1 & 100  & 15000 & (0, 0)  \\
LF-AmazonTitles-131k (\textsc{\texttt{FP8}}) & Distil-BERT  & 512  & 32  & 0.85 & 0.00001 &  0.05& 100 & 5000 & (0.01, 0.0001)  \\
\midrule
LF-WikiSeeAlso-320K (\textsc{\texttt{BF16}}) & Distil-RoBERTa   & 128  & 256  & 0.75  & 0.00002  & 0.08  & 100 & 5000  &  (0.01, 0.0001)   \\
LF-WikiSeeAlso-320K (\textsc{\texttt{FP8}}) & Distil-RoBERTa  & 128  & 256  & 0.75 & 0.00002 &  0.08& 100 & 5000  & (0.01, 0.0001) \\
\midrule
LF-AmazonTitles-1.3M (\textsc{\texttt{BF16}}) & Distil-BERT   & 512  & 32  & 0.65  & 0.000005  & 0.05  & 100  & 5000  & (0.0001, 0.0001)   \\
LF-AmazonTitles-1.3M (\textsc{\texttt{FP8}}) & Distil-BERT   & 512  & 32  & 0.6  & 0.000005 &  0.2 & 150  &15000  &  (0.01, 0.0001) \\
\midrule
LF-Paper2Keywords-8.6M (\textsc{\texttt{BF16}}) & Distil-BERT & 128 & 128 & 0.70 & 0.00002  & 0.05 & 12 & 5000 & (0.01, 0.0001)  \\
LF-Paper2Keywords-8.6M (\textsc{\texttt{FP8}}) & Distil-BERT & 128  & 128 & 0.70 &  0.00002&  0.05& 12 & 5000 & (0.01, 0.0001)  \\
\bottomrule
\end{tabular}%
}
\end{table*}

\begin{table}[ht]
\centering
\caption{Peak memory vs Latency (in terms of epoch time) for different chunk size when chunking classifier update is used with \texttt{BF16} on Amazon-3M dataset.}
\label{tab:chunk-size-comparison}
\resizebox{0.48\textwidth}{!}{%
\begin{tabular}{c|cc}
\toprule
\textbf{Chunk Size} & \textbf{Epoch Time (mm:ss)} & \textbf{Peak Mem (\SI{}{\gibi\byte})} \\
\midrule \midrule
1   & 13:22 & 14.74 \\
\midrule
2   & 12:20 & 14.40 \\
\midrule
4   & 12:12 & 12.22 \\
\midrule
8   & \textbf{11:09} & 11.13 \\
\midrule
16  & 11:23 & 10.59 \\
\midrule
32  & 12:39 & 10.32 \\
\midrule
64  & 14:19 & \textbf{10.20} \\
\midrule
128 & 19:44 & \textbf{10.20} \\
\bottomrule
\end{tabular}
}
\end{table}

% \section{Performance on the Commodity Hardware
% }
\section{Deployment on Commodity Hardware}
Although our main \texttt{Float8} experiments use H100 GPUs, we also demonstrate that the approach runs efficiently on commodity hardware, such as the GeForce RTX 4060 Ti. As shown in Table~\ref{tab:rxt4060}, the 4060 Ti uses slightly more memory, primarily due to the \texttt{torch.ao} package not yet fully supporting \texttt{Float8} on commodity hardware. As a result, we use a \texttt{BF16} encoder while still applying \texttt{Float8} to the XMC classifiers.

% We also evaluated the epoch time and memory consumption on commodity hardware, such as the GeForce RTX 4060. The results are shown in Table~\ref{tab:rxt4060}. Compared to the H100, it uses slightly more memory. This is because the \texttt{torch.ao} package does not yet fully support \texttt{Float8} on commodity hardware. As a result, we use a \texttt{BF16} encoder while still applying \texttt{Float8} to the XMC classifiers.

\section{Precision Recovery for Sensitive Applications}
For applications where recovering the last bit of accuracy is critical, two potential practical mitigation strategies that still operate within similar memory budgets are:

\begin{enumerate}
    \item \textbf{Post-hoc Classifier Refinement:} A simple approach is to fine-tune the classifier in higher precision on top of an ELMO-trained (low-precision) model using frozen encoder features. This allows a partial recovery of the lost precision while staying within a constrained memory budget by loading only subsets of labels at a time. This strategy introduces an additional training phase and hyperparameters to be tuned for the second stage.

    \item \textbf{Kahan summation for head labels:} To address accuracy drops without additional training stages, we also outline another approach that leverages label statistics inherent in XMC tasks. By exploiting the long-tailed label distribution, one can apply Kahan summation with \texttt{BF16} compensation only to the top-p\% most frequent labels. This approach selectively boosts precision@k, with minimal memory overhead, approximately 2×p\% (where p\% is memory for p\% label parameters in \texttt{Float8} ) more than the \texttt{Float8} baseline. Importantly, this strategy preserves end-to-end training and avoids the complexity of multi-stage pipelines. For example, 
 as shown in Table~\ref{tab:khan}, on AmazonTitles-1.3M with top 20\% head labels, this method achieves a competitive performance as Reene with a total classifier memory footprint of just 4.99 GB, still significantly below the \texttt{BF16} baseline (6.61 GB).
\end{enumerate}

\section{Tail Label Performance}\label{tail_label_perf} Table \ref{tab:psp-table} compares the propensity scored precision@k values, which is an indication of tail label performance. It would be interesting to explore the effect of tail label performance when going down in bit-width. Similar to the precision@k performance, our approach shows competitive performance with existing state-of-the-art XMC baselines, showcasing low precision training can be robust to tail labels. Performance of LF-AmazonTitles-131k is shown in Table \ref{tab:131k}.

% To further mitigate the performance drop on large label sets caused by quantization errors, we apply Kahan summation with \texttt{BF16} compensation specifically to the head labels. This approach narrows the accuracy gap between \texttt{BF16} and \texttt{FP8} by using higher-precision accumulation only for the most frequent labels. Although this method increases memory usage relative to pure FP8 classifiers, it remains significantly more memory-efficient than using full \texttt{BF16} classifiers.

% For example, as shown in Table~\ref{tab:khan}, on the LF-AmazonTitles-1.3M dataset, we apply a compensation vector in \texttt{BF16} for the top 20\% most frequent labels. This mitigation improves the P@1 score from 54.97 to 55.6, while still maintaining a lower memory footprint than the full \texttt{BF16} setup.

\section{Chunking Classifier Update: Latency vs Peak Memory}\label{chunking_ablation} Table \ref{tab:chunk-size-comparison} shows the latency (epoch time) vs peak GPU memory usage comparison for different chunk sizes. The training run is with ELMO BF16 data types for 3 million data size with batch size 128 on H100 GPU. We see chunking doesn't affect the latency until some point. In fact, we see the latency improves as the chunking increases from 1 to 8.

\section{Hyper-parameters and Implementation Details}\label{hyperandimplementations}
We detail the hyper-parameters of the \texttt{BF16} and \texttt{FP8} models in the table \ref{tab:hyper}.

\section{Low-Memory Dropout for Classfier}
To improve the classifier's robustness, we apply dropout \cite{pmlr-v28-wan13} to it. However, a limitation of traditional dropout is its requirement for a copy of the classifier's weights, doubling memory consumption. To address this, we implement dropout directly within the matrix multiplication process. After loading the classifier's weights from HBM into SRAM, we apply dropout to the loaded weights before performing matrix multiplication. This operation is handled within the Triton kernel, ensuring there is no additional memory overhead. We apply this for LF-AmazonTitles-1.3M.

% \section{Sensitivity of Low Precision Models}

\end{document}